%% file: main.tex
\documentclass[10pt,twocolumn,letterpaper]{article}

\usepackage{cvpr}
\usepackage{times}
\usepackage{epsfig}
\usepackage{graphicx}
\usepackage{amsmath}
\usepackage{amssymb}
\usepackage{amsthm}
\usepackage{bbm}
\usepackage{subfiles}
\usepackage{subfigure}

\usepackage[lined, boxed, ruled, commentsnumbered, noend]{algorithm2e}
\usepackage{xspace}
\usepackage{xpatch}

\usepackage[utf8]{inputenc} 
\usepackage[T1]{fontenc}    
\usepackage{url}            
\usepackage{booktabs}       
\usepackage{amsfonts}       
\usepackage{nicefrac}       
\usepackage{microtype}      

\input{macros}

\usepackage[pagebackref=true,breaklinks=true,colorlinks,bookmarks=false]{hyperref}

\cvprfinalcopy 

\def\cvprPaperID{1919} 

\ifcvprfinal\fi
\begin{document}

\title{Teaching Categories to Human Learners with Visual Explanations}

\author{Oisin Mac Aodha\hspace{20pt}Shihan Su\hspace{20pt}Yuxin Chen\hspace{20pt}Pietro Perona\hspace{20pt}Yisong Yue\\
California Institute of Technology
}

\maketitle

\begin{abstract}
We study the problem of computer-assisted teaching with explanations.  
Conventional approaches for machine teaching typically only provide feedback at the instance level, \eg, the category or label of the instance.  
However, it is intuitive that clear explanations from a knowledgeable teacher can significantly improve a student's ability to learn a new concept. 
To address these existing limitations, we propose a teaching framework that provides interpretable explanations as feedback and models how the learner incorporates this additional information.  
In the case of images, we show that we can automatically generate explanations that highlight the parts of the image that are responsible for the class label.
Experiments on human learners illustrate that, on average, participants achieve better test set performance on challenging categorization tasks when taught with our interpretable approach compared to existing methods. 
\end{abstract}


\subfile{sections/intro}

\subfile{sections/related}

\subfile{sections/method}

\subfile{sections/implementation}

\subfile{sections/experiments}

\subfile{sections/conclusion}

\small{
\noindent\textbf{Acknowledgments} We would like to thank Google for their gift to the Visipedia project, AWS Research Credits, Bloomberg, Northrop Grumman, and the Swiss NSF for their Early Mobility Postdoctoral Fellowship. Thanks also to Kareem Moussa for providing the OCT dataset and to Kun ho Kim for helping generate crowd embeddings.
}


{\small
\vspace{10pt}
\bibliographystyle{ieee}
\bibliography{main}
}

\clearpage

\subfile{sections/supplementary}

\end{document}

%% file: macros.tex
\newcommand{\tset}{T} 

\newcommand{\diff}{\text{diff}}

\newcommand{\dist}{\text{dist}}

\newcommand{\examples}{\mathcal{X}}
\newcommand{\ex}{x}

\newcommand{\STRICT}{\textsf{STRICT}\xspace}
\newcommand{\EXPLAIN}{\textsf{EXPLAIN}\xspace}
\newcommand{\EXPLAINCROWD}{\textsf{EXPLAIN\_CROWD}\xspace}
\newcommand{\RANDEXP}{\textsf{RAND\_EXP}\xspace}
\newcommand{\RANDIM}{\textsf{RAND\_IM}\xspace}

\newcommand{\err}{\textrm{err}}

\newcommand{\errate}{\textrm{err}}

\newcommand{\given}{\mid}

\newcommand{\tableref}[1]{Table~\ref{#1}}
\newcommand{\figref}[1]{Fig.~\ref{#1}}
\newcommand{\eqnref}[1]{\text{Eq.}~(\ref{#1})}

\DeclareMathOperator*{\argmax}{\arg\!\max}


%% file: sections/intro.tex
\vspace{-5pt}
\section{Introduction}
Computer-assisted teaching offers the promise of personalized curricula that are tailored to the ability level and interests of every individual. 
Providing open access to the kinds of high-quality teaching that is currently only available to a small percentage of the world's population has the potential to transform education.
To date, subject areas such as mathematics \cite{koedinger1997intelligent,canfield2001aleks} and language learning \cite{von2013duolingo} have benefited from automated teaching systems. 
However, the problem of how to best teach visual expertise in fine-grained domains such as medical diagnosis and species identification is comparatively less well explored.

In addition to the benefits to human learners from better teaching methods, automated systems could also take advantage of these improvements.
Access to expert time is often limited, and as a result, there is a need for better techniques to train crowd workers for image annotation tasks.
Once the workers have effectively learned the task they can provide higher quality labeled training data. 
This `closing of the loop' will enable us to take advantage of humans' ability to both generalize across different domains and cope with other nuisance factors such as pose and lighting changes. 

\begin{figure}[t]
    \centering
    \includegraphics[width=0.90\columnwidth]{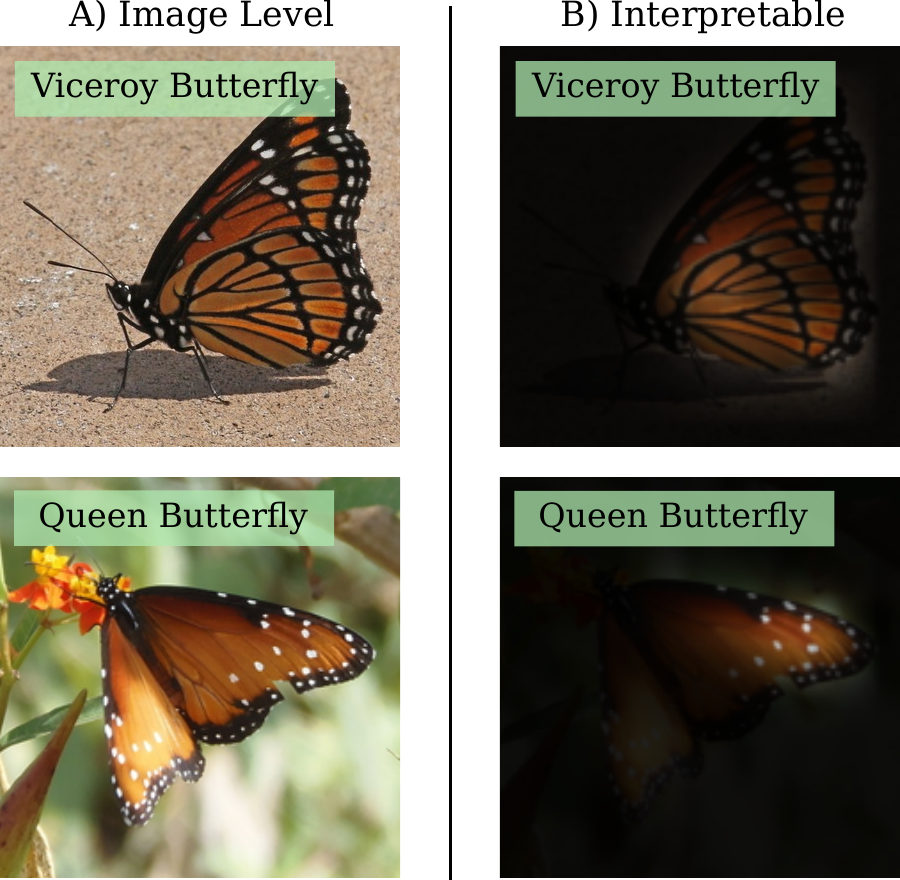}
    \caption{A) The majority of existing machine teaching algorithms for visual categories only give feedback to the learner in the form of the ground truth class label. 
    B) It is much more informative to display the discriminative regions to help them to determine the categories present in the images. Here, we see explanations from our system highlighting the blank band on the Viceroy Butterfly and the white spots on the Queen Butterfly's wings. These are field markings commonly used to identify both species.}
    \label{interp_feedback_fig}
\end{figure}

Existing approaches to teaching visual knowledge typically pose the problem as choosing the most informative subset of images to show from a much larger set of possible options. 
One of the major limitations of this existing work is that they only give very limited feedback to students in the form of the class label, \eg \cite{singla2014near,johns2015}. 
In Fig. \ref{interp_feedback_fig}  A) we see an example of this label only feedback. 
Providing only the ground truth class label is a very limited amount of feedback compared to the rich explanations that one may receive from a human teacher. 
A human teacher would likely teach the student the specific parts and attributes in the image that are most discriminative for that particular class, Fig.~\ref{interp_feedback_fig}~B).

Our hypothesis is that students that are taught with interpretable feedback will learn more effectively than those that only receive label feedback. 
Specifically, we propose a novel teaching algorithm that selects images that are both representative of the categories of interest and have explanations that can be easily interpreted. 
These explanations come in the form of feedback during teaching indicating the parts of the image that are important for successful classification. 
We show that it is possible to generate these explanations from existing labeled datasets, thus minimizing the need for additional annotations.
Through experiments on real human learners, we show that our joint selection of informative images and interpretable explanations results in better student learning and improved generalization at test time.

%% file: sections/related.tex
\section{Related Work}
\label{rel_work}

\subsection*{Machine Teaching}
The goal of machine teaching is the design of algorithms and systems that can teach humans efficiently and automatically. 
To date, a variety of different approaches have been explored for modeling the teaching of students from assuming perfect learners \cite{goldman1995complexity, zhu2013machine, liu2017iterative}, heuristic-based approaches \cite{basu2013teaching}, Bayesian models \cite{corbett1994knowledge,eaves2015tractable}, recurrent neural networks \cite{piech2015deep}, and reinforcement learning based methods \cite{rafferty2011faster,bak2016adaptive,whitehill2017approximately}.

While machine teaching has been successfully deployed in online tutoring systems that feature highly structured knowledge \eg mathematics, the teaching of challenging \emph{visual} concepts to human learners is less explored. 
\cite{singla2014near} teach binary visual classification tasks by modeling the student as stochastically switching between a set of different hypotheses during learning. 
Their model attempts to select the set of teaching examples offline that will best discount the incorrect hypotheses and guide the student towards the ground truth classification function.  
\cite{johns2015} propose an interactive approach, where the choice of future images to show is based on the individual’s past responses. 
However, they must update the model's parameters online for each user making it computationally difficult to scale to large numbers of simultaneous users in real-world settings.

The major limitation of these existing approaches is that the feedback they provide to the student is not fully informative. 
In both \cite{singla2014near} and \cite{johns2015} a student is shown a sequence of images and asked to estimate what object category from a finite list they believe to be present in each image. 
After they respond they are simply told what the correct answer is. 
They are not informed of the parts in the image that are discriminative for identifying that particular object, thus making the learning problem artificially hard for the student. 
To overcome this limitation, we propose a novel teaching algorithm that selects both images and provides interpretable explanations resulting in more understandable and efficient learning for the student. 
Complementary to our work, \cite{chenAiStats2018} recently introduced an explanation based teaching algorithm for binary tasks. Explanations are provided via pre-existing semantically meaningful features, but the interpretability of a given explanation is not modeled.  

\subsection*{Interpretable Models}
Using clear and understandable instructional material can dramatically improve a student's ability  to learn a new concept.  
It has been shown that highlighting informative regions on an image can help improve novice classification performance by guiding the student's attention \cite{grant2003,roads2016}.

In another example, when a human teacher is unavailable, the most common way novices learn species identification is by consulting expertly curated field guides. 
These field guides are typically books or apps that contain descriptive text and example images highlighting important features for classifying different species \eg \cite{peterson1980field}.
Attempts have been made to automate the creation of these guides using highlighted part annotations \cite{berg2013you}, automatic generation of image specific text descriptions \cite{hendricks2016generating}, or through gamification \cite{deng2016}. 
However, in addition to image level class labels, the majority of these approaches require the collection of \emph{expert} annotations in the form of text descriptions, anatomical part locations, or visual attributes which can be expensive and time consuming to obtain for very large image collections \cite{branson2010}. 
Furthermore, the efficacy of these annotations for teaching visual identification skills has not yet been evaluated on real human subjects. 
An alternative approach that requires less additional annotations is to learn human interpretable models from the raw data \eg \cite{ribeiro2016should, lakkaraju2016interpretable}.
In the context of computer vision, there is some evidence to suggest that the deep models commonly used for large-scale image classification tasks can be adapted to generate features that are semantically meaningful to humans \cite{zhou2016learning, zhang2017interpretable}.

Recently, \cite{pmlr-v54-poulis17a} outlined an approach for incorporating additional supervised data from users which they call `feature feedback'.
In addition to class level labels that are typically provided by human annotators when training supervised classifiers, they allow their annotators to provide information about the value of specific feature dimensions. 
In contrast, our approach instead gives explanations to the \emph{learner} about the importance of different image regions and models how they incorporate this information when updating their belief. 

We are concerned with selecting the set of teaching examples with associated interpretable explanations to best teach noisy human learners. 
While in practice these explanations can be generated with additional time-consuming human annotation, we show that it is possible to extract meaningful explanations using existing image level labels.

%% file: sections/method.tex
\vspace{-5pt}
\section{Interpretable Visual Teaching}
In this section, we outline our model for teaching categorization problems to human learners and present an efficient algorithm for selecting interpretable teaching examples. 
We discuss our approach in the context of the non-adaptive model of \cite{singla2014near}, but any visual teaching algorithm can benefit from interpretable explanations.

\subsection{Problem Setup}
We aim to teach a human learner a target classification function $h^*$ that maps from images $\mathcal{X}$ to their corresponding ground truth class labels $\mathcal{Y}$.
For instance, a single image $x \in \mathcal{X}$ could be a picture of a bird and the associated label $y$ could be the name of the species. 
However, we cannot directly impart the parameters of $h^*$ to the human learner, instead we must \emph{teach} them by showing them example images. 
Given a set of $n$ images, $\examples = \{\ex_1,\dots, \ex_n\}$, with associated labels from $C$ classes $\mathcal{Y} = \{y_1, \dotsc y_C \}$, our goal is to select an informative subset $\tset$, referred to as the teaching set, that will best convey the ground truth, \ie. the human teacher's classification function $h^*$.
Acknowledging that some images are more informative than others during learning, we do not want to waste effort teaching these unrepresentative images. 

When learning from a human teacher, or even from a textbook, a student not only receives feedback in the form of the correct answer but also an explanation that describes \emph{why} a given answer is correct.
In addition to the images and labels, we further assume that we (as the teacher) have access to an explanation, $e$, for each image $x$. 
In the case of images, these explanations could be heatmaps that highlight the informative regions for a particular class in a given image. 
In non-visual scenarios, this could be a piece of text describing the relationship between $x$ and $y$.

\subsection{Image Only Learner Model}
We adopt the stochastic \STRICT~algorithm of \cite{singla2014near} to model how learners adapt to the images shown by the teacher.
The model was originally proposed for teaching binary classification functions. 
Learners are modeled as carrying out a random walk in a finite hypothesis space $\mathcal{H}$. 
Each element of $\mathcal{H}$ is a function that maps from an image to a score $h: \mathcal{X} \mapsto \mathbb{R}$.
In the context of binary classification, where $y \in \{-1, 1\}$, the label predicted by a hypothesis $h$ for image $x$ is $\hat{y}^h = \text{sgn}(h(x))$, and the magnitude indicates the confidence that $h$ has in its prediction.
Concretely, an image $x$ may be represented by a feature vector and a hypothesis $h$ could be a linear classifier, $h(x) = w_h^{\intercal}x$, with weights $w_h$.
To ensure that it is possible to teach the learner, we assume that $\mathcal{H}$ also contains the ground truth hypothesis $h^*$. 

At the beginning of teaching, the learner randomly picks a hypothesis $ h \in \mathcal{H}$ according to the prior distribution $P(h)$. 
During teaching, she will be presented with a sequence of images along with the correct class label. 
After receiving a new image, the learner will stick to her current hypothesis if the ground truth label is consistent with the prediction of the hypothesis.
Otherwise, she randomly switches to a new $ h \in \mathcal{H}$ according to her current posterior over the hypotheses $P(h\mid \tset)$, where $\tset$ is the set of teaching images and ground truth labels seen so far.
When updating her posterior $P(h\mid \tset)$ in light of the new information, hypotheses that disagree with the true labels of the images shown by the teacher are less likely to be selected 
\begin{align}
P(h\mid \tset) \propto P(h) \prod\limits_{\substack {x_t\in{} \tset\\y_t \neq \hat{y}^h_t}} P(y_t\mid h, x_t),
\label{equation:posterior}
\end{align}
where 
$\hat{y}^h_t$ is the class label predicted by hypothesis $h$ for image $x_t$.
We model how much the prediction of hypothesis $h$ agrees with the correct label for image $x_t$ using
\begin{equation}
    P(y_t\mid h, x_t) = \frac{1}{1 + \exp(-\alpha h(x_t)y_t)},
\end{equation}
where  $\alpha > 0$ represents how consistently a learner responds according to her currently adopted hypothesis. 
In the extreme when $\alpha = \infty$, hypotheses that are inconsistent with the ground truth label are immediately discarded.
This unrealistic setting represents the perfect learner~\cite{goldman1995complexity}.

\subsection{Interpretable Feedback}
The feedback presented to the learner in the \STRICT model is comprised of only the true, ground truth, label of a teaching example. The learner is tasked with learning the mapping between the image and the ground truth label themselves as no other explanations are provided \ie, they must determine which regions or parts in the image are responsible for the given class label. 
In real-world teaching scenarios, the teacher often has access to much richer information which can be further utilized to accelerate the teaching process.
Our approach, \EXPLAIN, gives feedback to the learner in the form of \emph{explanations} and models how they incorporate this information when updating their belief.

With a slight abuse of notation, we extend $\tset$ to denote the set of labeled teaching images along with their explanations. Our updated model of the learner introduces two additional discount terms that account for the interpretability of an explanation and the representativeness of an image
\begin{equation}
P(h\mid \tset) \propto P(h) \prod\limits_{\substack {x_t\in{} \tset\\y_t \neq \hat{y}^h_t}} P(y_t\mid h, x_t) \prod\limits_{x_t\in \tset}\left(E(e_t) D(x_t)\right).
\end{equation}

\subsubsection*{Modeling Explanations}
Our first new term favors explanations for images that are clear and easy for the learner to understand
\begin{equation}
   E(e_t) = \frac{1}{1 + \exp(-\beta\; \diff(e_t))}.
   \label{eqn:explanation}
\end{equation}
Here, $\diff(e_t)$ represents how difficult a given explanation $e_t$ is for image $x_t$, where large values indicate challenging explanations.
Intuitively, the learner is more confident in discounting inconsistent hypotheses if presented with easier-to-understand explanations.
This information could be crowdsourced, but later we describe a method by which both the explanations and their interpretability can be automatically generated from image sets with ground truth labels.

Note that the quality of an explanation is not measured in the same way as image difficulty.
Image difficulty in \STRICT is implicitly encoded by the image's location in the feature space. 
Images that are close to decision boundaries are assumed to be more ambiguous to the learner compared to those that are far from decision boundaries.
However, two images of equal difficulty may have very different explanations and our goal is to bias the selection of teaching images towards those that are easier to understand.

\subsubsection*{Modeling Representativeness}
\begin{figure}[t]
\centering
\includegraphics[width=0.48\textwidth]{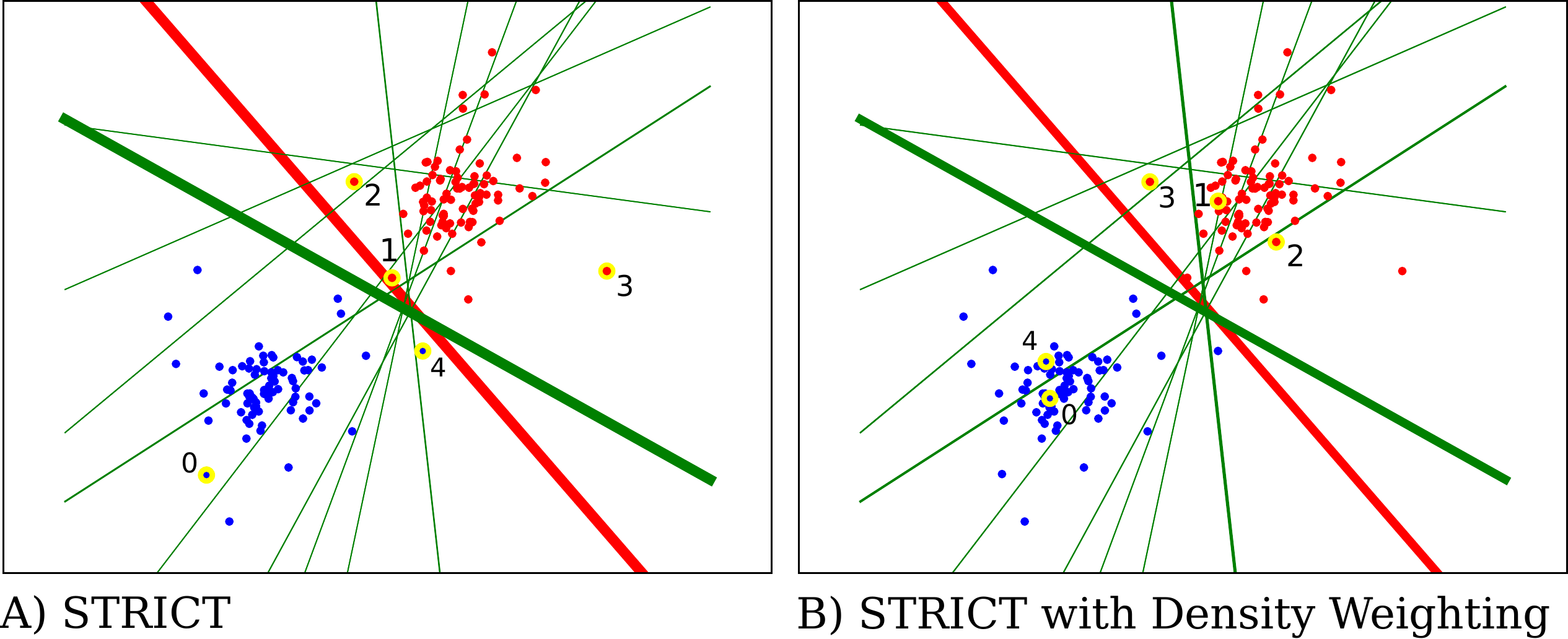}
\caption{In A) we see that \STRICT is prone to selecting outliers. Here, the numbers represent the order in which examples were selected during teaching. B) By favoring instances that are \emph{representative} of others  we select teaching examples from more dense regions of the feature space. Here, the lines represent different hypotheses, with the red one being the teachers hypothesis. The thickness of the lines represent the probability associated with the hypothesis after showing five teaching examples to the learner.} 
\label{fig:density}
\end{figure}

During teaching, we would like to select \emph{representative} teaching images so that the concept conveyed to the learner can be easily generalized to the remaining, unseen, images. 
In practice however, we observe that the linear hypothesis space used by \STRICT has a tendency to select outlier images (Fig.~\ref{fig:density}). 
When selecting the teaching set, \STRICT attempts to greedily optimize the expected reduction in error for the hypothesis set and as a result can end up selecting images that may be optimal for reducing error but not necessarily informative for the learner. 
To overcome this limitation, inspired by approaches in active learning \cite{settles2012active}, we include an additional discount factor that favors representative examples
\begin{equation}
   D(x_t) = \frac{1}{1 + \exp(-\gamma\; \dist(x_t))}.
\end{equation}
Here, $\dist(x_t)$ encodes how dissimilar an image $x_t$ is to other images. Again, this could be derived directly from crowdsourcing but a simple alternative is to use the density in the existing feature space
\begin{equation}
   \dist(x_t) = \frac{1}{N}\sum_{n=1}^{N}{||x_t -x_n||_2^2}.
\end{equation}
In practice, we compute the mean distance for each example to all other examples in the same class.
The teacher still aims to select informative examples, but this allows us to ensure that they are also representative.
Setting both $\beta  = \gamma = \infty$ results in the existing \STRICT model. 

\subsection{Multiclass Teaching}
Many real word teaching problems feature multiple different categories of interest. 
As presented, \STRICT is limited to the binary class setting.
One approach for handling multiple classes is to change the hypothesis space so that each individual hypothesis is a multiclass classifier \eg, a softmax classifier where individual hypotheses are one-versus-all classifiers.
However, generating such a multiclass hypothesis space is challenging.
For instance, if one has access to a set of linear classifiers $\mathcal{H}^L$ \eg crowdsourced from a system like \cite{welinder2010multidimensional}, one could assume that each multiclass hypothesis is made up of a combination of $C$ of these linear classifiers, where $C$ is the number of classes.
This results in $|\mathcal{H}^L|! / (|\mathcal{H}^L|-C)!$ possible combinations, which becomes prohibitively large to exhaustively cover for even small numbers of hypotheses and classes.

Instead, we model a separate posterior $P_c(\mathcal{H}\mid \tset)$ for each of the classes.
This allows us to reuse the same set of hypotheses across the different classes. 
It naturally enables us to model the situation where a learner can accurately group images based on their visual similarity but are unable to associate the correct class label to the groups as they have yet to learn the correct mapping.
When teaching binary concepts, this corresponds to the single posterior modeling of the existing \STRICT model.

\subsection{Teaching Algorithm}
Given our model of the learner, we now describe how we select the teaching set $\tset$. 
The choice of $\tset$ is based on our desire to direct the learner towards a distribution over the possible hypotheses that results in the smallest number of mistakes as possible. In our multiclass setting with $C$ classes, we use a one-versus-all scheme for each class, and maintain $C$ different posterior distributions over the hypotheses $\mathcal{H}$. Intuitively, in order to drive down the learner's error probability in predicting multiclass labels, it is sufficient to make sure that the learner performs well in each of the $C$ binary classification tasks. For any class $c$, we define the error of a one-vs-all hypothesis $h$ over all possible images as
\begin{align*}
  \err_c(h) = \frac{|{x: (\hat{y}^h \neq y_c \wedge y=y_c) \vee (\hat{y}^h = y_c \wedge y\neq y_c)}|}{| {\mathcal X} |}.
\end{align*}
This is the fraction of images that hypothesis $h$ disagrees with the ground truth when predicting the label $y_c$. After receiving the teaching set $T$, the expected error of the learner for class $c$ is defined as
\begin{align}
  \mathbb{E}[\err_c(h)\mid \tset] = \sum \limits_{h \in \mathcal{H}}P_c(h\given T)\err_c(h).
\end{align}
For our multiclass teaching problem, we use the combined error probability, \ie., $\frac1C\sum_c \mathbb{E}[\err_c(h)\mid \tset]$
as a proxy for the expected error of the learner. Given a teaching budget $b$, we would like to find a teaching set $T^*$, such that upon observing the images, their labels, and associated explanations the learner would achieve the maximal reduction in expected error. Formally, let
\begin{align}
  R(\tset) &= \frac1C\sum_c\left(\mathbb{E}[\errate_c(h)] - \mathbb{E}[\errate_c(h)\given \tset] \right)\nonumber \\
           &=\frac1C\sum_{c \in \mathcal{C}} \sum \limits_{h \in \mathcal{H}}(P_c(h) - P_c(h\given \tset))\err_c(h)\label{eqn:objective}
\end{align}
be the expected reduction in the combined error term. We aim to find
\begin{align}
  \tset^* = \argmax_{|\tset| \leq b} R(T).
  \label{eqn:problem}
\end{align}
Instead of directly optimizing this challenging combinatorial problem we use the greedy submodular approach outlined in~\cite{singla2014near}. 
We start with an empty teaching set $\tset=\emptyset$ and greedily add a single image at a time.
The selection of the next teaching image to show amounts to choosing the example $x$ from the unseen set ($x\notin \tset$),
\begin{equation}
  x_{t} = \argmax_{x} R(\tset \cup \{x\}).
  \label{eqn:final_selection}
\end{equation}

%% file: sections/implementation.tex
\section{Implementation Details}
In this section we outline how we automatically generate explanations, our hypothesis space, and how to efficiently optimize the teaching objective.

\subsection{Image Explanations}
For each image, we require an explanation $e$ that tells the learner why an image $x$ has a given label $y$.
How to best generate explanations is still an open question \cite{DoshiKim2017Interpretability}.
One, time-consuming, way to acquire these explanations would be to ask an expert to manually label the informative regions in each image. 
Alternatively, they could be crowdsourced, but this may result in very noisy explanations \eg, \cite{deng2016}.
Instead, we propose to use the ground truth class label provided with each image to automatically generate visual explanations. 

We exploit the fact that modern Convolutional Neural Networks (CNNs) used for image classification often produce semantically meaningful features.   
We use the Class Activation Mapping approach of \cite{zhou2016learning} to automatically generate explanations for each image, but other existing methods for generating explanations can also be easily used with our model.
For each pixel location $j$ we computed the weighted sum of the output of the final convolutional layer from a CNN that has been trained on the input data,
\begin{equation}
 e(j) = \sum_k w_c^k f^k_{j}(x) + b_c.
\end{equation}
Here, $f^k_{j}(x)$ denotes the feature value at pixel location $j$ and output channel $k$ for the CNN $f$.
The weight values and biases, $w_c$ and $b_c$, from the final fully connected layer associated with the ground truth class are used to weight each feature map. 
Finally, we normalize the explanations so that each spans the range $[0, 1]$.

We use the entropy of the explanation 
as a proxy for the difficulty the user may have in interpreting it,
\begin{equation}
 \diff(e) = -\frac{1}{J}\sum_{j}e(j)\log(e(j)),
\end{equation}
where $J$ is the number of pixels in the explanation.
This captures our preference for localized and discrete highlighted regions whose values are either $0$ or $1$. 
In practice, weight regularization applied during the training of the CNN ensures that we do not have explanations that are uniformly high, and the classification training objective ensures that there are some non-zero entries in the feature maps to correctly predict the class label.
So there is no bias towards classes that have low entropy on average, we subtract the mean difficulty for each class (making sure the final difficulties are $>0$).

We use a ResNet18~\cite{he2016deep} trained from scratch on each dataset as our backbone CNN and extract features from the output of the second residual block, resulting in a downsampling factor of $8$.
To this, we append an additional convolutional layer with $64$ filters of size $3\times3$, and a final fully connected layer with the number of classes as output.
We finetune the entire model on $128\times128$ image crops with a batch size of $64$ for $60$ epochs using Adam \cite{kingma2014adam}. 
During training, we use random flips, crops, and color augmentations.
We start with an initial learning rate of $0.0002$ and decay by a factor of $10$ after $30$ epochs.
To generate the final explanation map $e$ for each $144\times144$ input image $x$, we extract the explanation for the center $128\times128$ crop and pad and resize it to input resolution.

\subsection{Hypotheses Space Generation}
We require access to a representative hypothesis space $\mathcal{H}$ to perform teaching. 
This space should span the different possible linear classifiers that the learners may be using. 
It represents the different biases that they may have in relation to the teaching task at hand. 
However, generating such a space for multiple fine-grained categories is not trivial.
As an alternative to generating embeddings from crowd annotations \eg \cite{welinder2010multidimensional}, we propose to use the features from our finetuned CNN from the previous section. 
Extracting the $64$ dimensional penultimate feature vector from our modified ResNet18 gives us a representation that encodes visual similarity.
While this is unlikely to be the same representation used by our learners it has the advantage that it can be generated with no additional annotation cost. 

We generate a set of candidate linear classifiers using the CNN features by first clustering the features in each class into $2$ subsets and training a linear SVM to separate each from the rest of the data.
We also train a linear SVM to separate each class from the other classes, representing the best possible hypothesis $h^*$. 
Next, we group each pair of classes and train additional linear models to separate these from the remaining classes.
We add all these linear classifiers to the hypothesis set and add an additional random linear classifier to bring the total number of hypotheses to $100$.
Each dataset is split into $80\%$ training and $20\%$ test sets.
As in \cite{singla2014near}, images in the training set that are not possible to correctly classify with the optimal hypothesis $h^*$ are removed.

\subsection{Efficient Optimization}
Optimizing \eqnref{eqn:final_selection} involves searching over all unseen images in $\mathcal{X}$ and measuring their reduction in error for each hypothesis in $\mathcal{H}$ across all classes. 
With the aid of some pre-computation, we can reduce the updates to simple matrix operations.
For simplicity, we present this in terms of the binary \STRICT model \cite{singla2014near}, but the same formulation holds for our multiclass \EXPLAIN algorithm by including the additional discount factors and classes.

First, we note that as the prior term in \eqnref{eqn:objective} is constant at each time step and can be removed without affecting the selection of $x_{t}$.
Second, we can re-factor the posterior into the current posterior multiplied by the update
\begin{equation}
R(T \cup \{x\}) = - \sum \limits_{h \in \mathcal{H}}\err(h) P(h\given T) \delta(h, x),
\end{equation}
where $\delta(h, x) = P(y|h, x)$ if $y \neq \hat{y}^h$ and otherwise it is set to $1$.
In matrix notation, this can be rewritten as 
\begin{equation}
\boldsymbol{r} = -(\boldsymbol{e}\circ\boldsymbol{p}_i) \boldsymbol{L},
\end{equation}
where $\circ$ is the element-wise multiplication.
$\boldsymbol{e}$ is a vector of size $1\times |\mathcal{H}|$ that contains the error associated with each hypothesis, and can be computed once offline.
$\boldsymbol{p}_i$ is a vector of size $1\times |\mathcal{H}|$ that represents the current posterior over the  hypotheses, which we will update after each time step $i$.
$\boldsymbol{L}$ is a $|\mathcal{H}|\times |\mathcal{X}|$ matrix that can be computed once offline. 
Each entry in $\boldsymbol{L}$ encodes the confidence each $h$ places in each example $x$ 
\begin{equation}
\boldsymbol{L}_{h,x} =
\begin{cases}
  1, & \text{if}\ y = \hat{y}^h \\
  P(y\given h, x), & \text{otherwise}.
\end{cases}
\end{equation}
Computing the next teaching example, $x_t$ simply amounts to choosing the maximum entry in the $1\times |\mathcal{X}|$ error reduction vector $\boldsymbol{r}$, taking care to only select from the examples not currently in the teaching set. 
We then update the posterior using 
\begin{equation}
\boldsymbol{p}_{i+1} = \boldsymbol{p}_i \circ \boldsymbol{l}_{t}, 
\end{equation}
where $\boldsymbol{l}_{t} = \boldsymbol{L}_{:, t}$ is the column vector from $\boldsymbol{L}$ associated with the selected teaching example $x_t$.
For the multiclass setting, we simply maintain a separate $\boldsymbol{e}^c$, $\boldsymbol{L}^c$, and $\boldsymbol{p}^c$ for each class and choose the next teaching example that results in the biggest reduction across all classes.

%% file: sections/experiments.tex
\section{Experiments}

\begin{figure*}[t]
\centering
\includegraphics[width=1.0\textwidth]{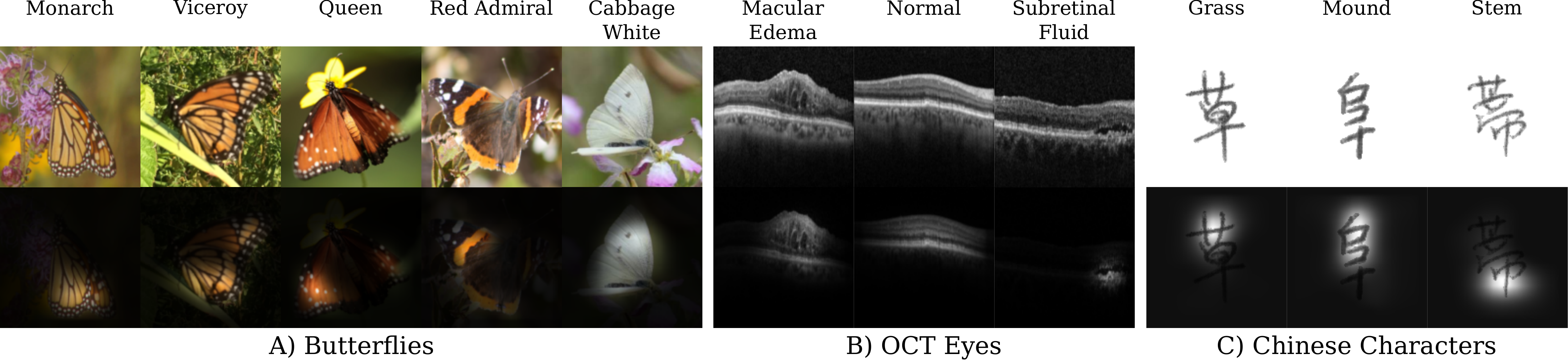}
\caption{Here we see example images from our three datasets with their corresponding, automatically generated, visual explanations below.} 
\label{fig:datasets}
\vspace{-10pt}
\end{figure*}

\subsection{Datasets}
There are existing datasets with explanations in the form of annotated visual attributes that could be used to teach visual categories to human learners.
However, we found that in many cases, \eg, \cite{wah2011caltech}, the provided attributes were too coarse-grained (\eg `forehead color') to be useful for teaching highly similar categories.
Other alternatives were too noisy to be informative \eg \cite{deng2016}, but present an interesting avenue for future human-in-the-loop explanation generation.
Instead, we selected three diverse datasets to span a range of different teaching use cases.
Example images and explanations from each dataset can be seen in \figref{fig:datasets}. 

\noindent{\bf Butterflies}
Our first dataset represents the teaching of naturalists who wish to identify different species of plants and animals in the wild. 
It contains images of five different species of butterflies captured in a large variety of real-world situations with varying image quality from the iNaturalist dataset~\cite{van2017inaturalist}.
Two of the species, Monarch and Viceroy, are very similar in appearance and the third, Queen, looks the same from underneath.
The final two species, the Red Admiral and the Cabbage White are distinct in appearance. 
We extracted a bounding box around each butterfly and manually discarded images of caterpillars.
In total, the dataset contains 2,224 images, close to uniformly distributed across each of the species. 

\noindent{\bf OCT Eyes} 
Our second dataset mimics the teaching of a trainee ophthalmologist that is tasked with learning how to identify different retinal diseases in images. 
It consists of image slices of the retina obtained through optical coherence tomography (OCT). 
OCT is a non-invasive imaging technique that uses light waves to take cross-sectional images of the retina. 
It is commonly used in ophthalmology, where on the order of 30 million OCT scans are performed around the world annually \cite{swanson2011ophthalmic}. 
Here, the learner's goal is to classify each of the images into one of three classes: Normal, contains Macular Edema, or contains Subretinal Fluid.
Diabetic Macular Edema is one of the leading causes of vision loss for people with diabetes \cite{romero2011managing}. 
Many of the examples are challenging and require subtle inspection to correctly identify the condition.
The dataset contains 1,125 images with ground truth annotations provided by retina specialists.

\noindent{\bf Chinese Characters} 
Finally, we use a dataset of three different Chinese characters extracted from \cite{liu2011casia}, that was also used in \cite{johns2015}.
Here, we are exploring the scenario of a non-native Chinese speaker attempting to learn to identify new characters. 
Each of the 717 images is a different handwritten example of one of the following three characters: Grass, Mound, or Stem. 
The images vary in difficulty as there is a large variety in the handwriting quality and style for the different individuals that contributed to the dataset.

\subsection{Experimental Setup}
We conducted experiments with real human participants on the crowdsourcing platform Mechanical Turk using the same methodology as \cite{singla2014near, johns2015}. 
Crowd workers, \ie learners, were randomly assigned to a dataset and baseline teaching strategy. 
On average we received $40$ participants per strategy and dataset combination. 
We assumed that the learners were motivated to perform well but acknowledge that this might not always be the case. 
First, the learners were shown a brief tutorial illustrating how our web-based teaching interface worked. 
When teaching began, they were presented with a sequence of images, shown one at a time, selected by the assigned teaching algorithm.
After viewing each image in turn, they were then asked to select from a list of options indicating the class they believed the image to contain.
Depending on the assigned teaching strategy, they were given feedback either in the form of the correct class label, or the correct class label and its corresponding visual explanation.
The visual explanations were displayed on top of the input images, alternating between the input image and the explanation every 0.5 seconds. 
Upon receiving feedback, learners had to wait for a minimum of 2 seconds before they could proceed to the next teaching image. 
For each dataset we showed $20$ teaching images, followed by $20$ randomly selected testing images where no feedback was provided during testing. 
This random sequence of test images was different for each learner. 
Longer teaching sequences would likely lead to better test performance but there is a trade off between the learner's attention span and their accuracy. 
We shuffled the order of the response buttons depicting the class labels for each learner before teaching began to ensure that there was no bias towards a particular button position.

We compared our full \EXPLAIN model to three baseline algorithms: 1) \RANDIM random selection of images, 2) \RANDEXP random selection with visual explanations, and 3) our multiclass version of \STRICT \cite{singla2014near}. 
Note, this last baseline did not include density weighting or explanations.
For both \EXPLAIN and \STRICT we generated a fixed teaching sequence once offline and used it for all learners with the same hypothesis space for both.
We set the noise level of the learners to $\alpha = 0.5$, and the explanation and representativeness parameters to $\beta = \gamma = 1.0$.

\begin{figure*}[t]
    \centering
    \subfigure[Butterflies]{  
        \centering
        \includegraphics[trim={0px 0px 0px 20px},clip, width=.22\textwidth]{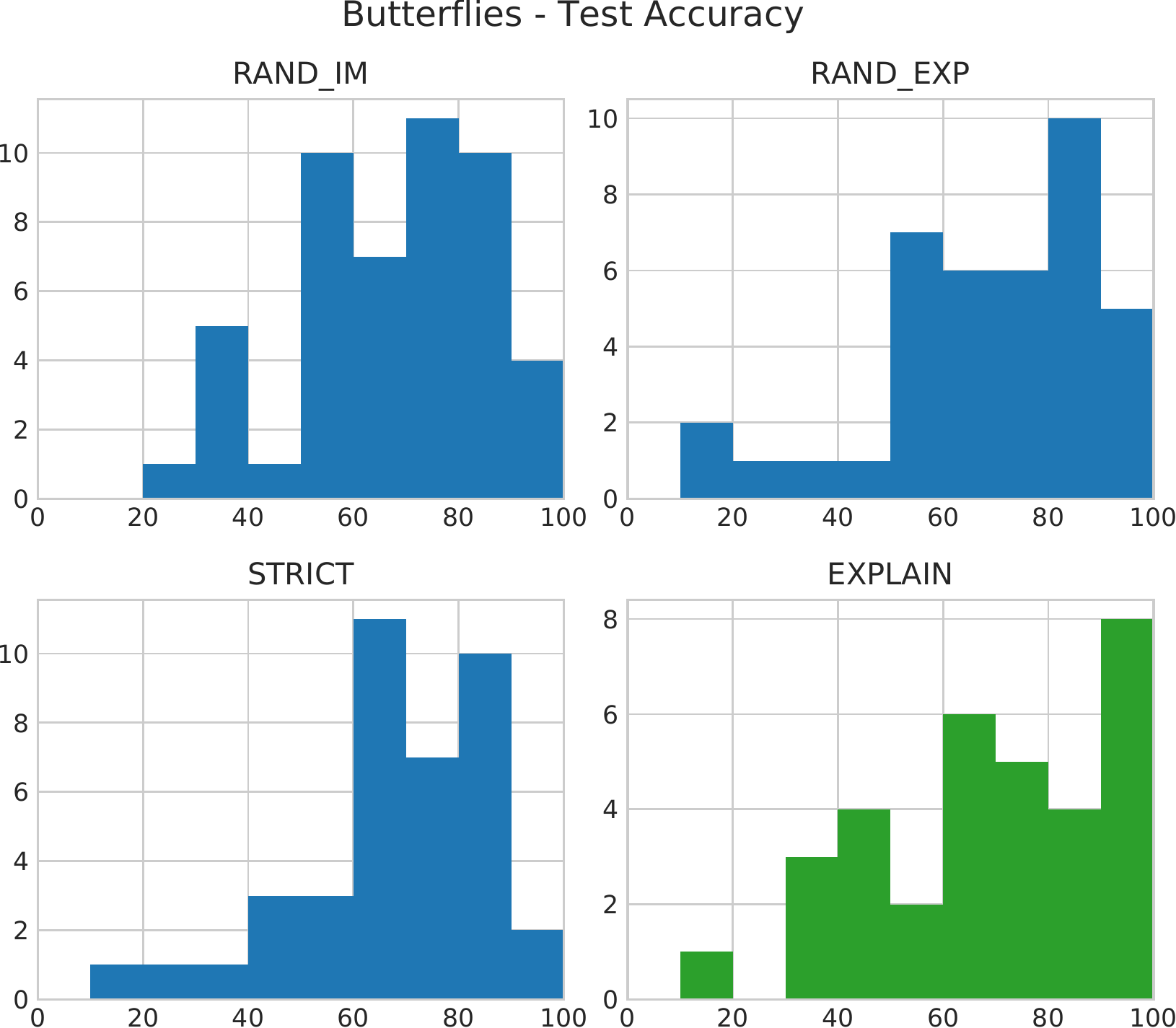}
    }~
    \subfigure[OCT Eyes]{
        \centering
        \includegraphics[trim={0px 0px 0px 20px},clip, width=.22\textwidth]{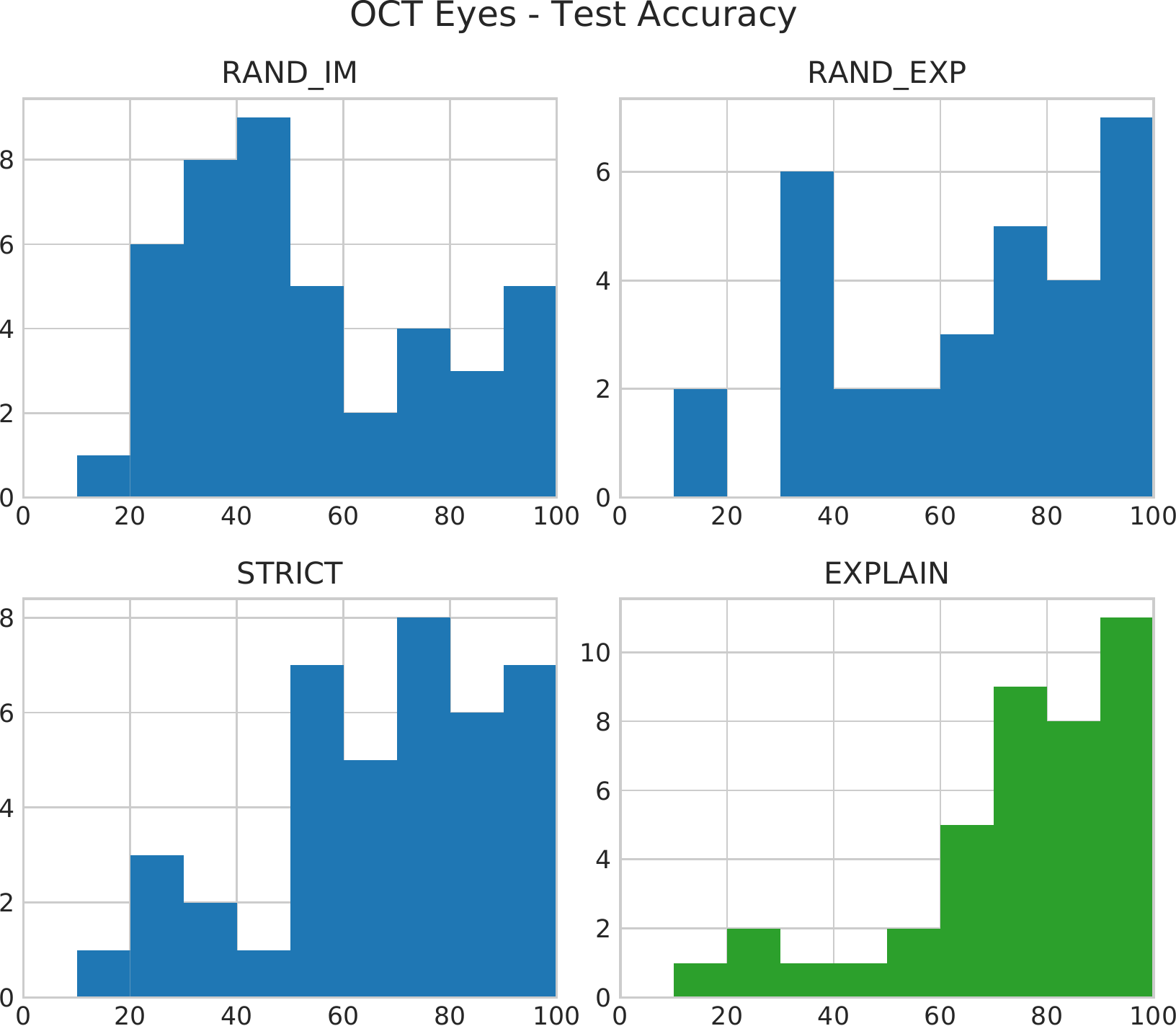}
    }~
     \subfigure[Chinese Characters]{
       \centering
        \includegraphics[trim={0px 0px 0px 20px},clip, width=.32\textwidth]{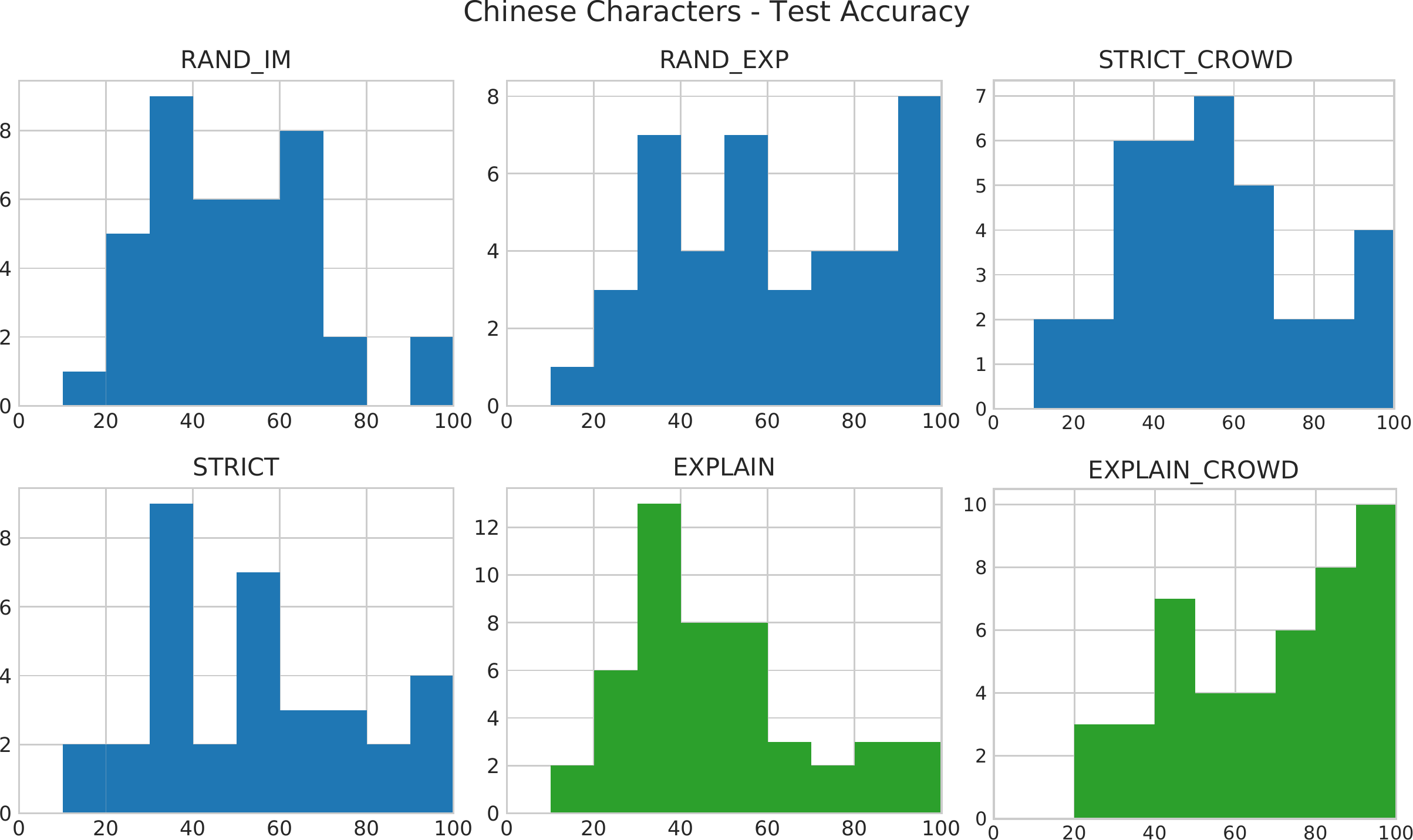}
    }~
    \subfigure[OCT Eyes Confusion]{
        \centering
        \includegraphics[trim={0px 0px 0px 0px},clip, width=.20\textwidth]{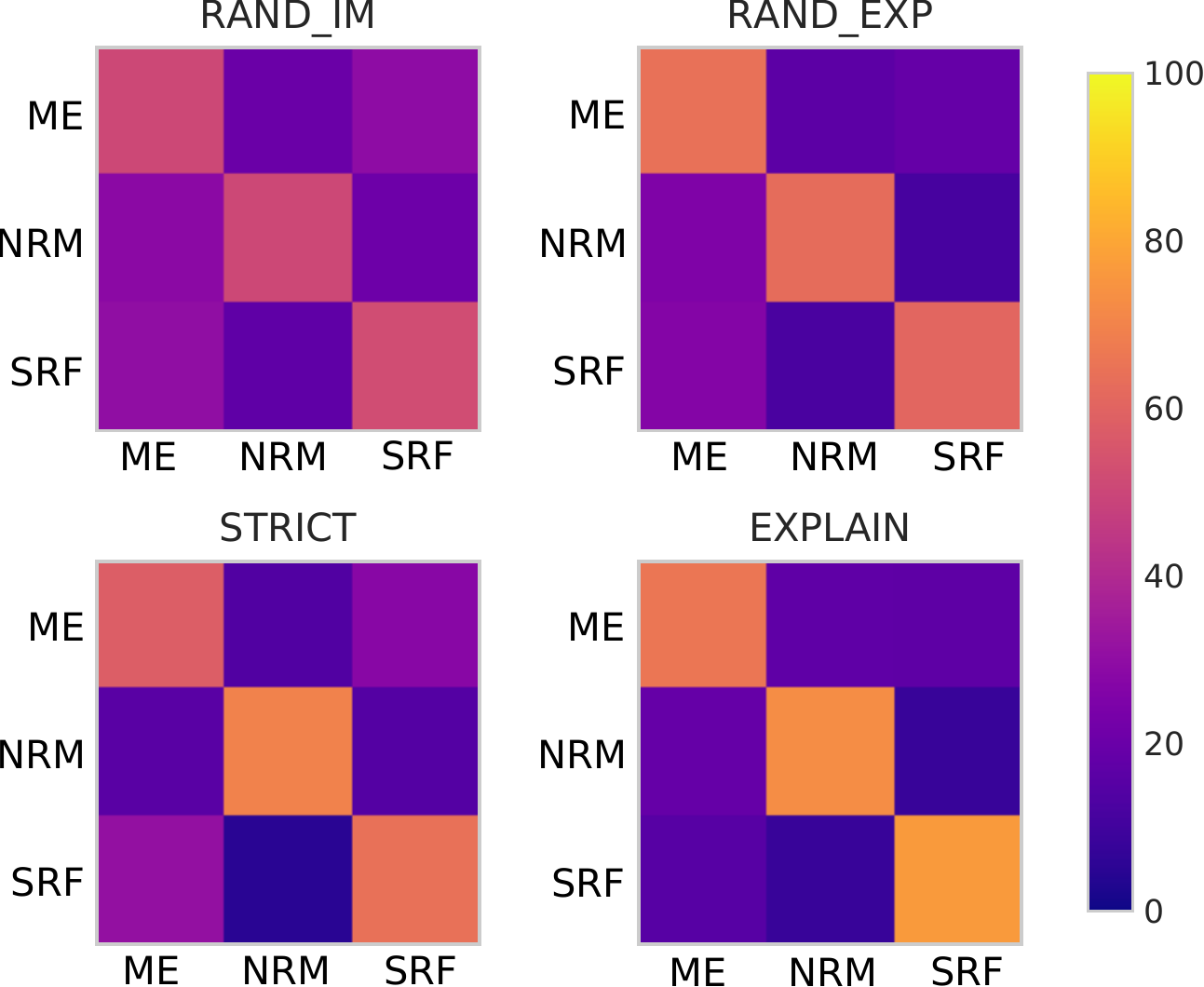}
    }
    \caption{Test time classification performance for human learners. A) - C) Show learners binned by test time accuracy. The horizontal axes represent average test scores, where larger numbers indicate higher accuracy. The vertical axes are the number of learners. D) Test time learner confusion matrices for the OCT Eyes dataset. We see that both explanation based strategies result in smaller off-diagonal entries.}
    \label{fig:results}
    \vspace{-10pt}
\end{figure*}

\subsection{Results}
In \tableref{tab:results_summary} we report the average test time accuracy for each learner for the different baseline teaching strategies and datasets.
\figref{fig:results} A) - C) displays the histograms of these accuracies illustrating that strategies that include explanations tend to do better overall.

\begin{table}[t]
\scriptsize
\centering
\begin{tabular}{|l|l|l|l|l|}
\hline 
                   &\RANDIM&\RANDEXP&\STRICT&\EXPLAIN \\ \hline\hline
Butterflies        & 65.20  & 67.31          & 65.00  & {\bf 68.33} \\
OCT Eyes           & 51.05  & 62.58          & 64.63  & {\bf 72.38} \\
Chinese            & 47.05  & {\bf 58.90}    & 51.91  & 46.35       \\
Chinese - Crowd      &        &                & 53.06  & {\bf 65.44} \\\hline
\end{tabular}
\caption{Average test time accuracies for Mechanical Turk learners across the three different test datasets.
Compared to the standard image only baseline \RANDIM, the inclusion of our explanations in \RANDEXP results in better test time performance across all datasets.
Our \EXPLAIN model results in superior learners in two of the three datasets with CNN generated hypothesis spaces.}
\label{tab:results_summary}
\vspace{-10pt}
\end{table}

For the Butterflies dataset we see that a greater percentage of learners get high scores when taught with \EXPLAIN, see \figref{fig:results} A). 
This is a challenging dataset, and we observe that many learners remain confused between the three similar species by the end of teaching, perhaps necessitating longer teaching sequences. 

The OCT Eyes images are likely to be the most unfamiliar to our learners compared to the other datasets.
However, the images are well aligned and do not contain large out of plane view point changes or confusing background texture. 
This enables us to generate high-quality explanations that localize the characteristic morphologies for the different retinal diseases. 
For the \RANDIM baseline in \figref{fig:results} B) the difference in learner performance can be explained by the images selected and their intrinsic motivation.
In \figref{fig:results} D) we see the average test time confusion matrices across all learners for the OCT Eye dataset.
It is clear that \EXPLAIN results in less cross-category confusion \ie smaller values in the off-diagonal entries.
We see that learners tend to make fewer mistakes when identifying Subretinal Fluid as it is relatively distinct, while there is more confusion between Macular Edema and Normal. 

\begin{figure}[h]
    \centering
    \includegraphics[width=0.75\columnwidth]{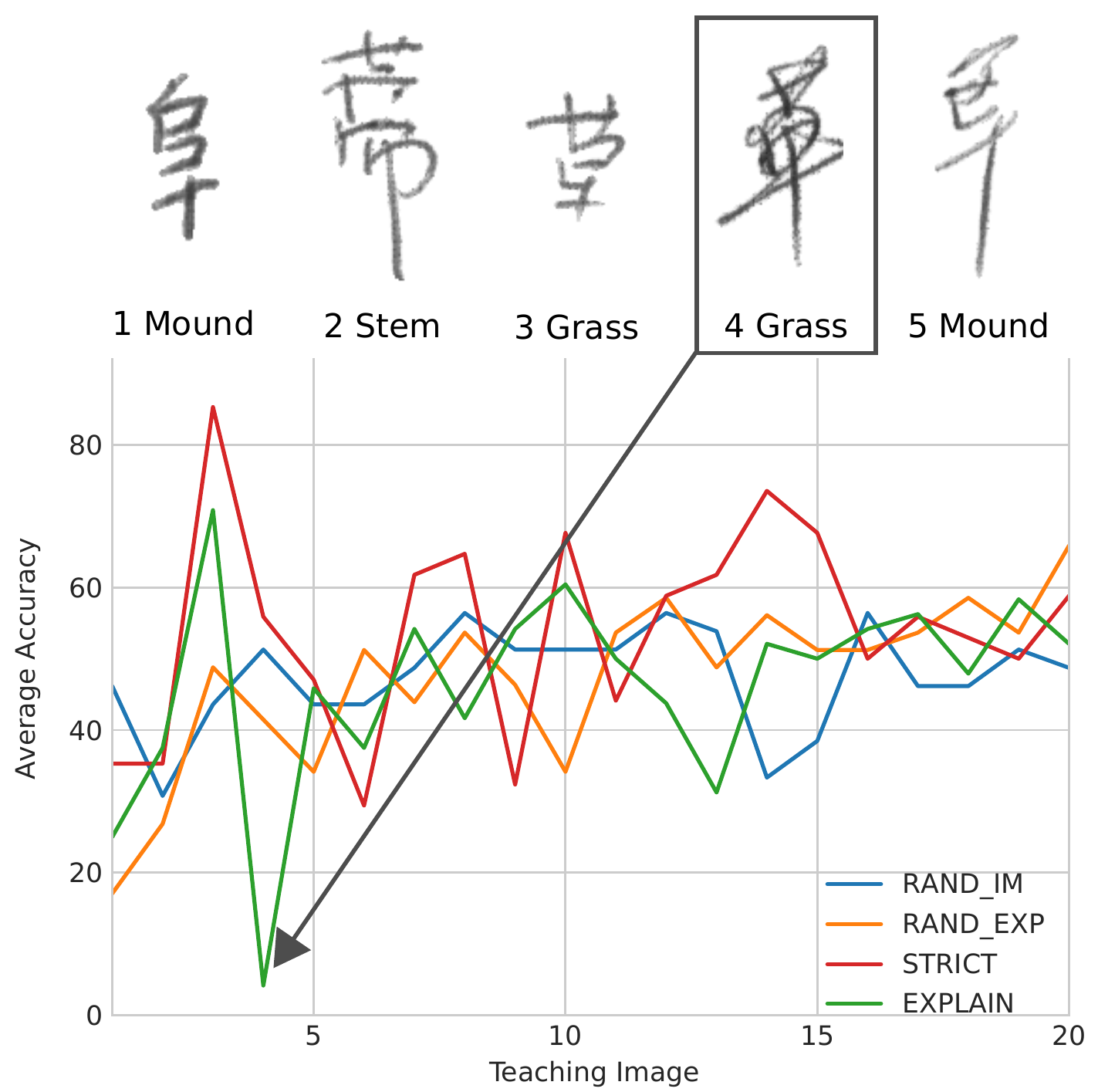}
    \caption{Average accuracy during teaching for the Chinese Characters dataset using the CNN generated hypothesis space, where random guessing is $33\%$. For all strategies, we observe a general improvement in learners' ability over time. Above we see the first five teaching images selected by \EXPLAIN. The fourth image is a difficult instance, resulting in the vast majority of learners guessing the wrong category.}
    \label{fig:chinese_fail}
    \vspace{-10pt}
\end{figure}

The Chinese Characters dataset represents an interesting failure case for \EXPLAIN when using the CNN generated hypothesis space, see \figref{fig:results} C). 
In \figref{fig:chinese_fail} we see the average performance during teaching for all learners and the first five teaching images selected by \EXPLAIN. 
\EXPLAIN selects a particularly difficult image for the fourth teaching example.
It happens to be the same class as the previous image, but is visually very different.
From inspecting the test time confusion matrix we see that this early difficult example potentially biases learners as they end up with lower performance for the \emph{Grass} class. 
The random based strategies have the advantage of generating different teaching sequences for each learner by uniformly sampling the input space thus introducing some additional robustness.
In \tableref{tab:results_summary} we see that \RANDEXP performs well on this dataset, indicating that the performance dip for \EXPLAIN may be a result of the automatically generated hypothesis space or the explanation interpretability scores, rather than the quality of the explanations themselves.
To test this hypothesis, we assigned manual interpretability scores and generated a separate embedding space more closely aligned with human notions of similarity by soliciting pairwise similarity estimates  on Mechanical Turk to construct a new embedding space (Chinese -  Crowd)\footnote{Full details are provided in the supplementary material.}.
Teaching with this embedding space (\EXPLAINCROWD) results in the best test time performance overall, see the last row in \tableref{tab:results_summary}.

%% file: sections/conclusion.tex
\section{Conclusion}
\label{conclusion}
We introduced \EXPLAIN, an algorithm for teaching multiple visual categories to human learners with interpretable feedback.
Interpretable feedback can be applied to any existing visual teaching algorithm.
It provides cues to the learner during teaching in the form of visual explanations and explicitly models how they incorporate this information when updating their belief. 
This enables us to generate more informative teaching sequences resulting in improved test time performance \ie, better generalization from the learners on unseen images. 
Experiments featuring real human participants show that our approach is superior to existing methods that only provide weaker class label feedback.
In future, we plan to investigate extending our approach to the interactive teaching setting, where the model updates online based on the learner's responses \eg, \cite{singla2013actively,du2011active,johns2015}. 

%% file: sections/supplementary.tex
\onecolumn
\noindent{\Huge Supplementary Material}
\section{Crowd Generated Hypothesis Space}
\begin{figure*}[h!]
    \centering
    \subfigure[CNN Embedding]{  
        \centering
        \includegraphics[trim={10px 0px 10px 25px},clip,width=.32\textwidth]{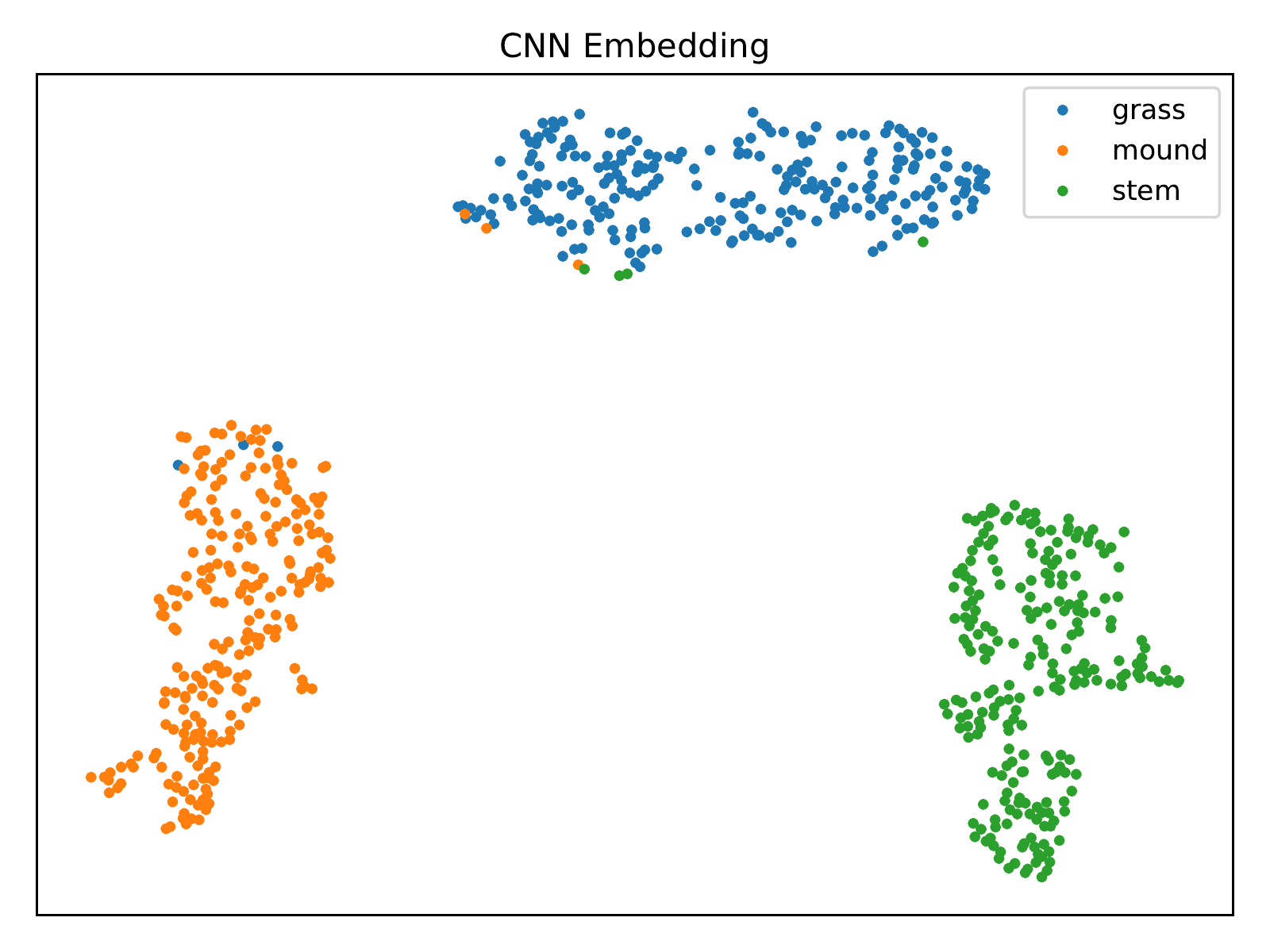}
    }~
    \subfigure[Crowd Embedding]{
        \centering
        \includegraphics[trim={10px 0px 10px 25px},clip,width=.32\textwidth]{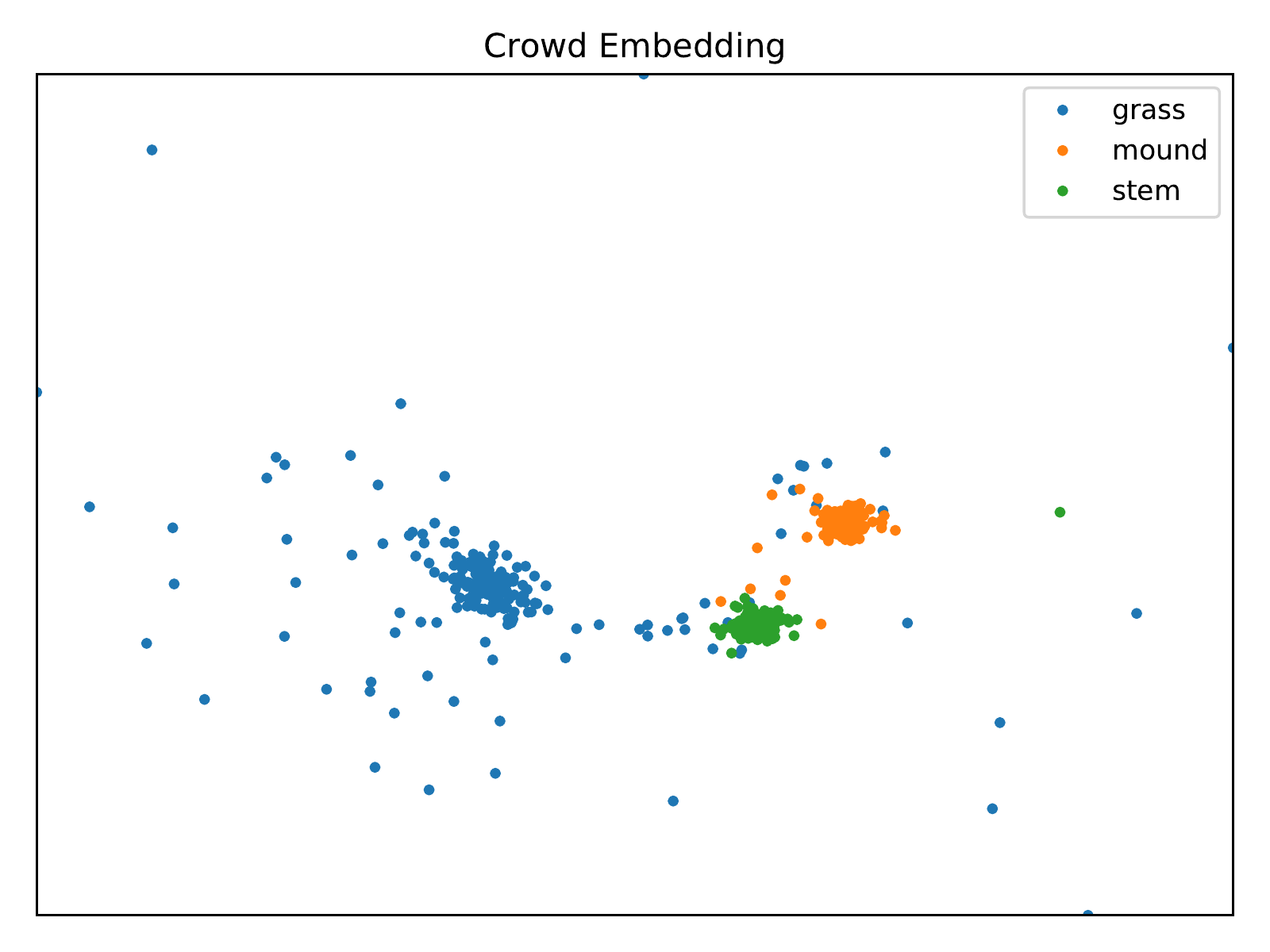}
    }~
     \subfigure[Explanation Difficulty]{
       \centering
        \includegraphics[trim={10px 0px 10px 25px},clip,width=.32\textwidth]{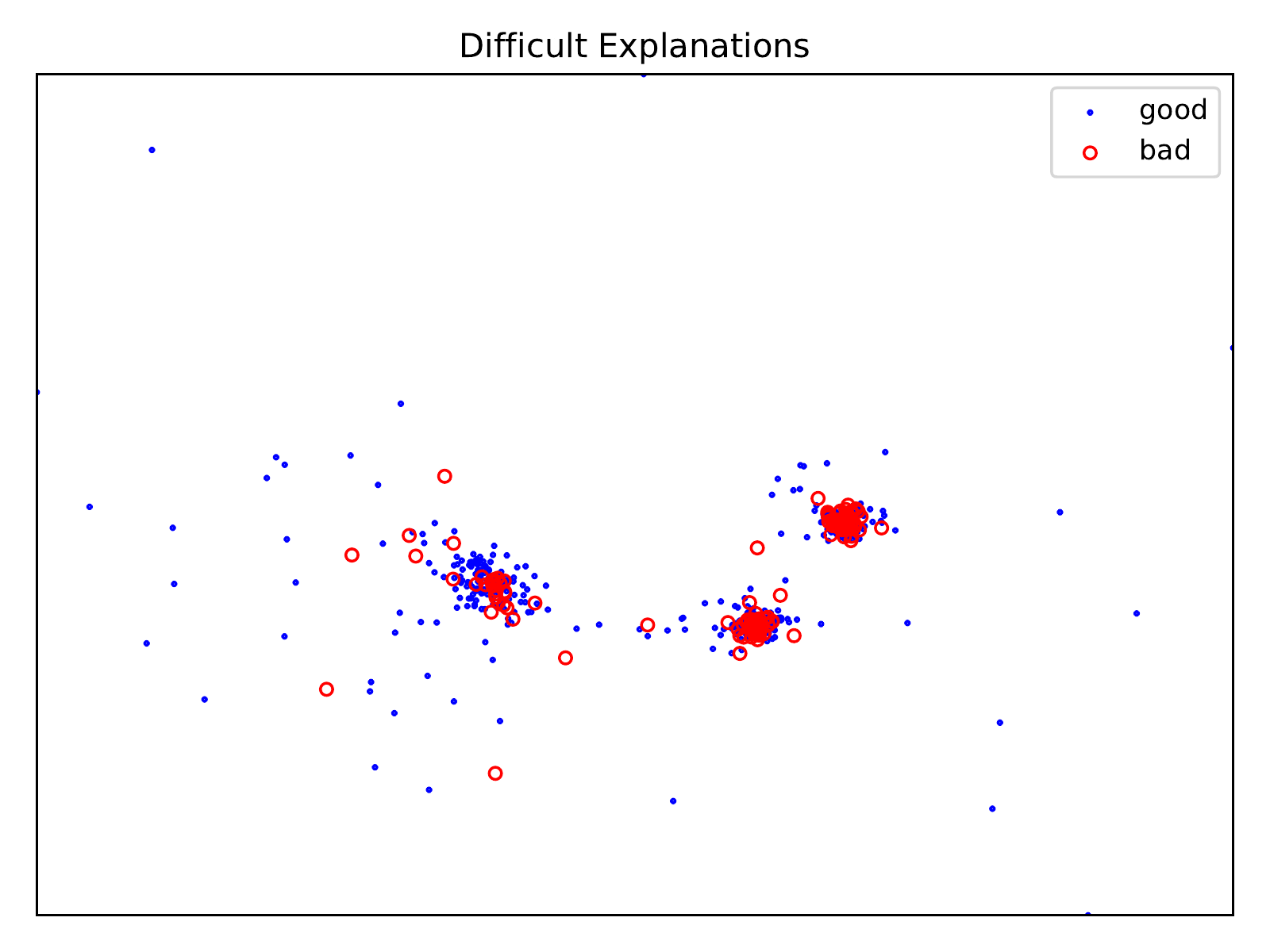}
    }
    \caption{CNN versus crowd embeddings for the Chinese Characters dataset. A) 2D embedding generated with t-SNE~\cite{maaten2008visualizing} of our 64D CNN features. B) Embedding generated with t-STE~\cite{van2012stochastic} from crowd triplet based similarity constraints. C) The red circles depict manually filtered `bad' explanations. We observe that these bad explanations are not necessarily correlated with difficult examples \ie, examples that are located between classes.}
    \label{fig:chinese_embed}
\end{figure*}

In the main paper, our \EXPLAIN algorithm using the CNN generated hypothesis space performed poorly on the Chinese Characters dataset.
To address this problem, we collected an alternative data embedding that better matches the learner's notion of similarity. 
Using a similar grid-based interface to \cite{gomes2011crowdclustering,kim2018context} with a $4\times6$ image grid, workers on Mechanical Turk were asked to group images of the Chinese Characters based on their appearance.
They were not given any information regarding the ground truth class labels. 
We then converted these groupings into a set of triplet constraints \ie image A is closer to B, compared to C, if A and B were placed in the same cluster by a given worker and C was in a different cluster.
In total, this resulted in 414,160 triplets from 51 workers. 
We used t-STE~\cite{van2012stochastic} to generate a 5D embedding of the images which was then used to construct our hypothesis space. 
This 5D space is used as an alternative to the CNN features from the main paper.
For the hypothesis generation, we used the same strategy outlined in the main paper, along with the same values for the hyperparameters $\alpha$, $\beta$, and $\gamma$, the same duration of teaching and testing, and the same train and test split.
For visualization purposes, a 2D embedding of the images can be viewed in \figref{fig:chinese_embed}.

Another potential weakness of our fully automatic approach is the automatic estimation of difficulty for each explanation $e$. 
In the main paper, we proposed using the entropy of the explanation image as a proxy for the difficulty, $\diff(e)$, that a learner may have in interpreting it.
As an alternative, here, we manually inspected each of the 717 explanations for the Chinese Characters dataset and labeled 589 of them as being `good' explanations and the rest as `bad'.
An explanation was deemed `bad' if it did not clearly highlight the discriminative regions for that particular class \eg the first image in \figref{fig:chinese_examples} A).
In practice, we set a difficulty value of $0$ for `good' explanations and $1000$ for `bad' ones. 
\EXPLAIN can still potentially select images with bad explanations, but the high cost discourages their selection. 

We ran an additional experiment on Mechanical Turk with the crowd generated hypothesis space and expert curated explanation difficulties. 
For comparison, we also measured the performance of \STRICT using the same hypothesis space.
As \RANDIM and \RANDEXP do not depend on the hypothesis space the results from the main paper for these strategies can be directly compared.  
From a total of 81 learners, where each individual was randomly assigned to one of the two strategies, we observed a test time average accuracy of 65.44\% for \EXPLAIN and 53.06\% for \STRICT.
This is an improvement for \EXPLAIN compared to the 46.35\% test accuracy achieved with the hypothesis space generated from CNN features.
Again, we qualitatively observe that \STRICT has a tendency to sometimes select outliers, but it still performs better than random image selection.

\section{Sample Images and Explanations}
\begin{figure*}
\centering

    \subfigure[Monarch]{  
        \centering
        \includegraphics[width=0.95\textwidth]{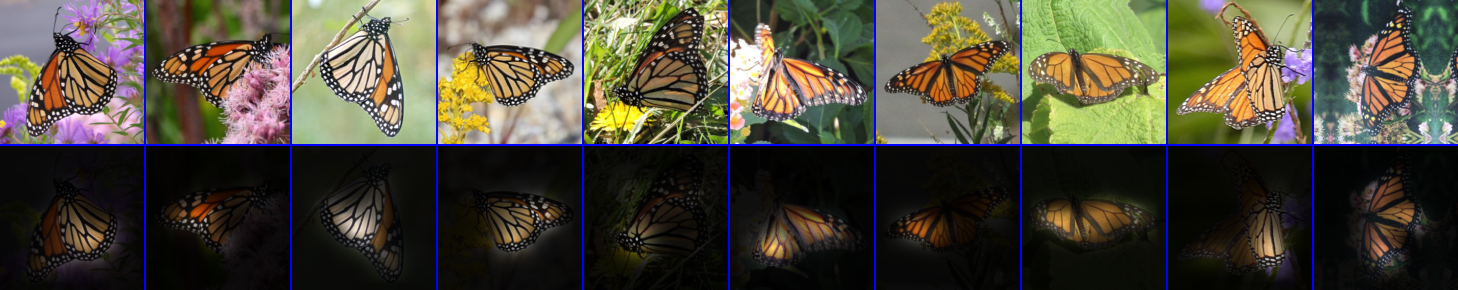}
    }\\
    \subfigure[Viceroy]{
        \centering
        \includegraphics[width=0.95\textwidth]{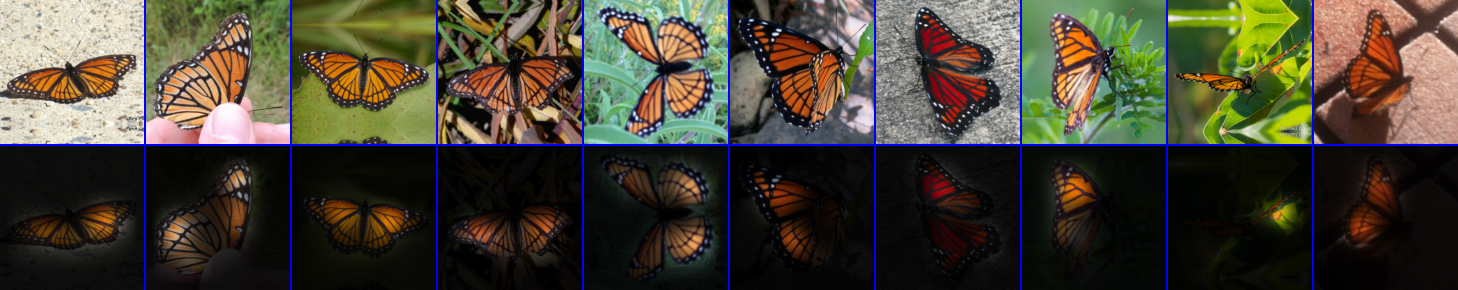}
    }\\
     \subfigure[Queen]{
       \centering
        \includegraphics[width=0.95\textwidth]{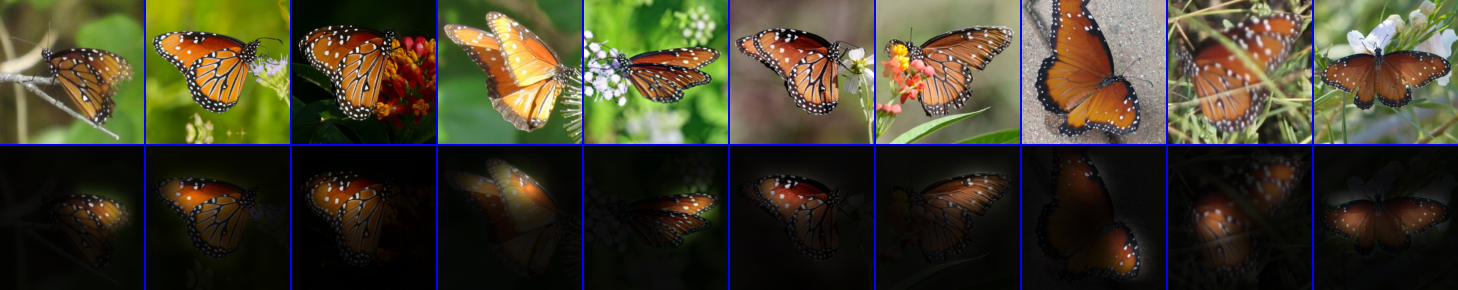}
    }\\
        \subfigure[Red Admiral]{
        \centering
        \includegraphics[width=0.95\textwidth]{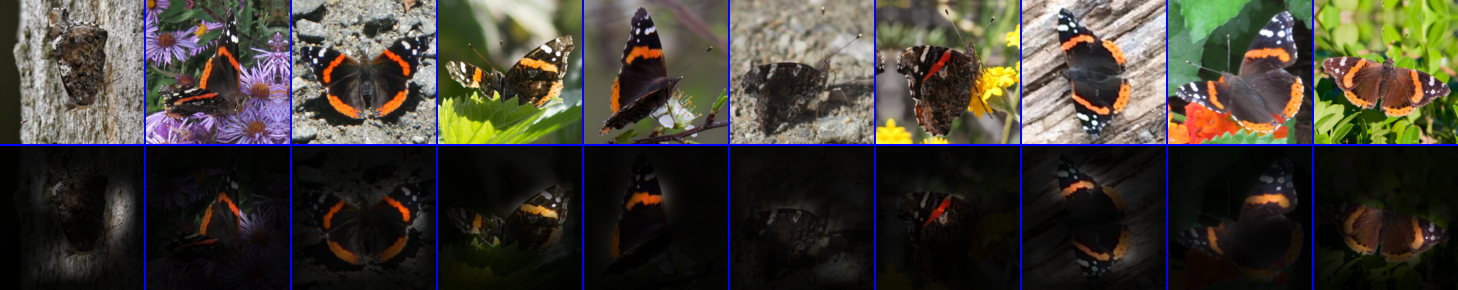}
    }\\
    \subfigure[Cabbage White]{
        \centering
        \includegraphics[width=0.95\textwidth]{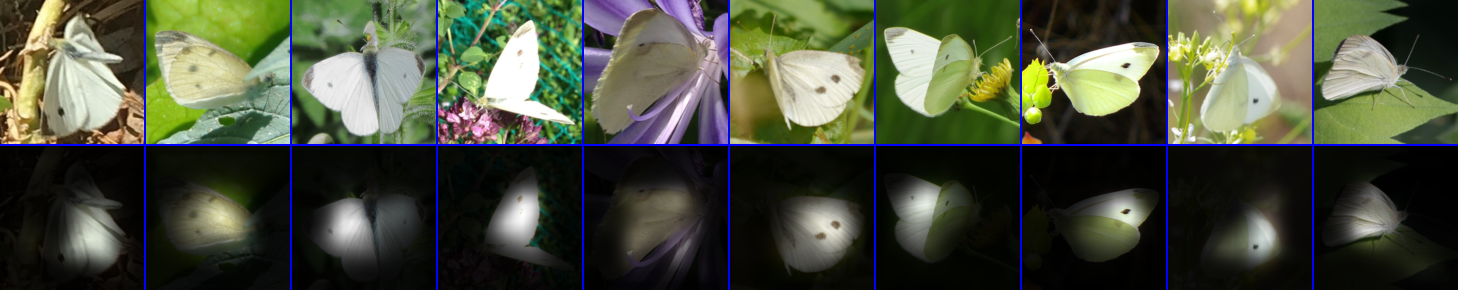}
    }

\caption{Randomly selected images and their corresponding explanations for each class from the Butterflies dataset. The second last Viceroy image depicts a poor explanation as the butterfly is in an unusual pose.}
\label{fig:butterflies}
\end{figure*}

\begin{figure*}
\centering

    \subfigure[Macular Edema]{  
        \centering
        \includegraphics[width=0.95\textwidth]{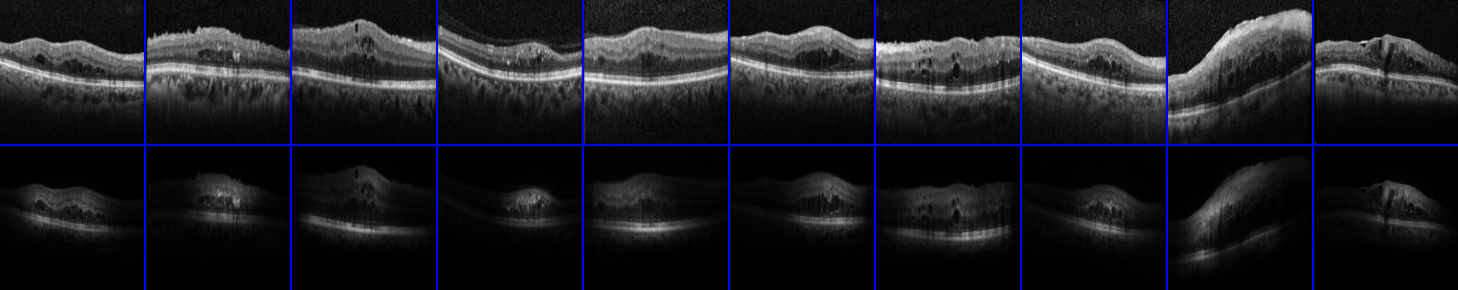}
    }\\
    \subfigure[Normal]{
        \centering
        \includegraphics[width=0.95\textwidth]{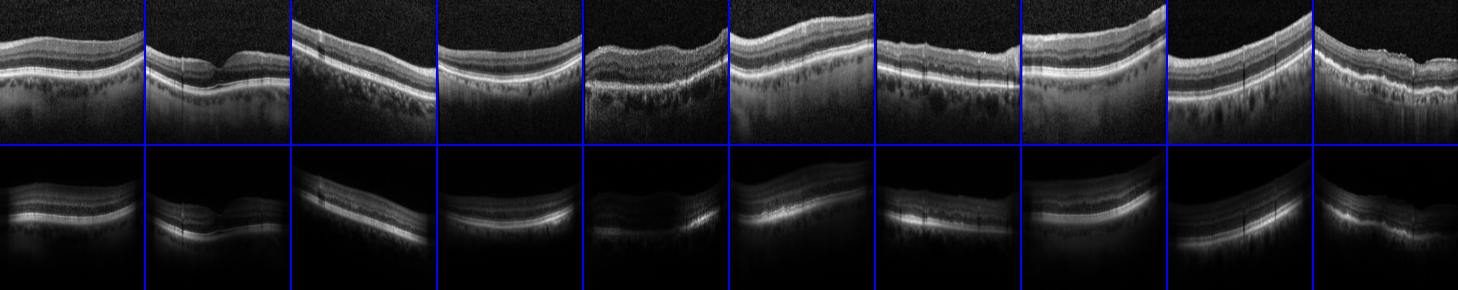}
    }\\
     \subfigure[Subretinal Fluid]{
       \centering
        \includegraphics[width=0.95\textwidth]{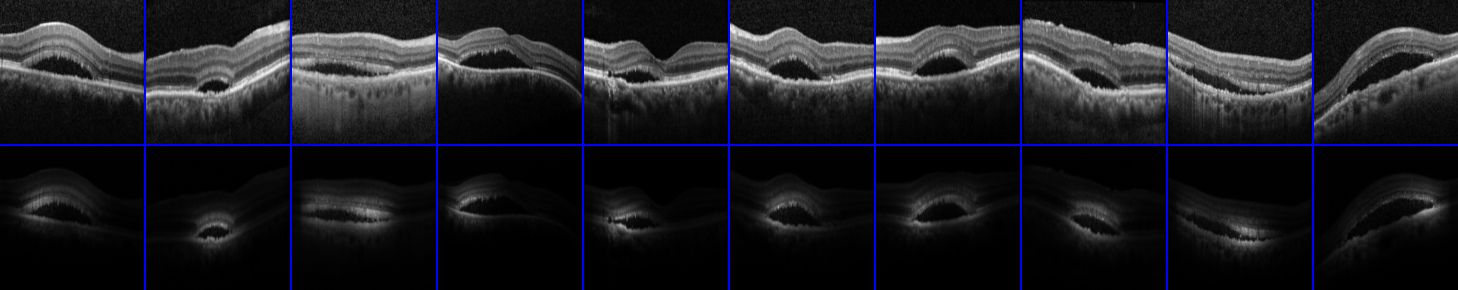}
    }

\caption{Randomly selected images and their corresponding explanations for each class from the OCT Eyes dataset.}
\label{fig:oct}
\end{figure*}

\begin{figure*}
\centering

    \subfigure[Grass]{  
        \centering
        \includegraphics[width=0.95\textwidth]{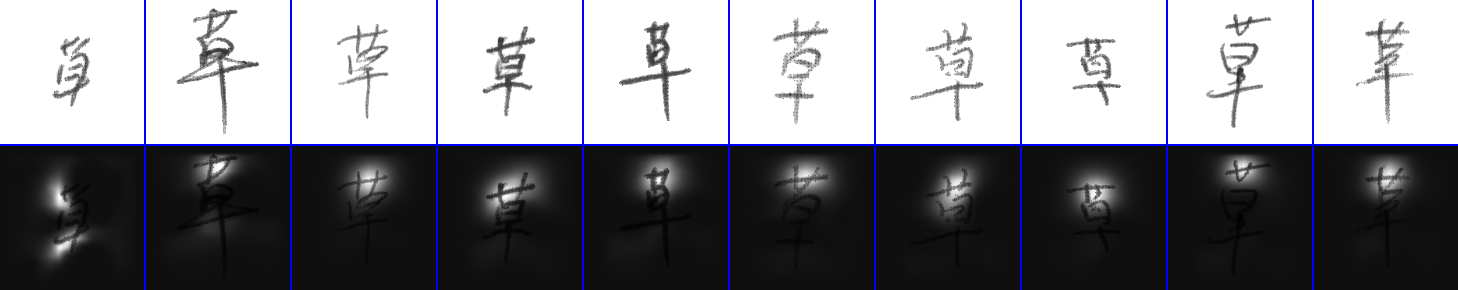}
    }\\
    \subfigure[Mound]{
        \centering
        \includegraphics[width=0.95\textwidth]{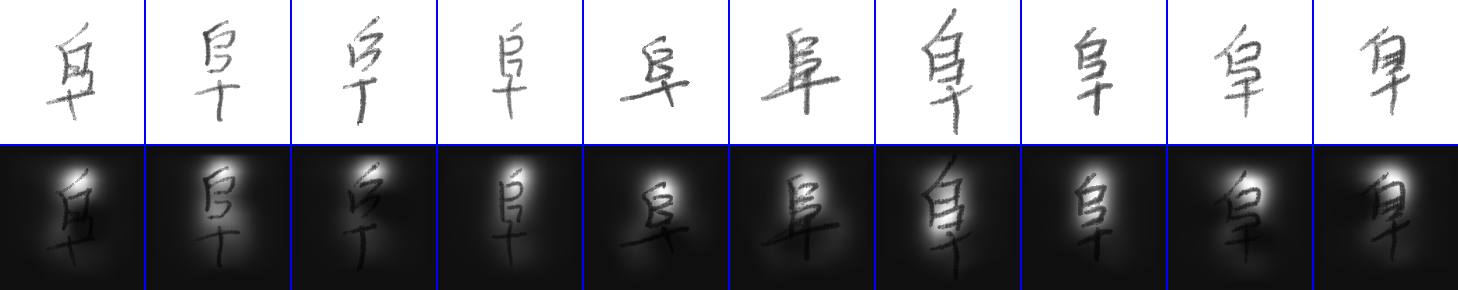}
    }\\
     \subfigure[Stem]{
       \centering
        \includegraphics[width=0.95\textwidth]{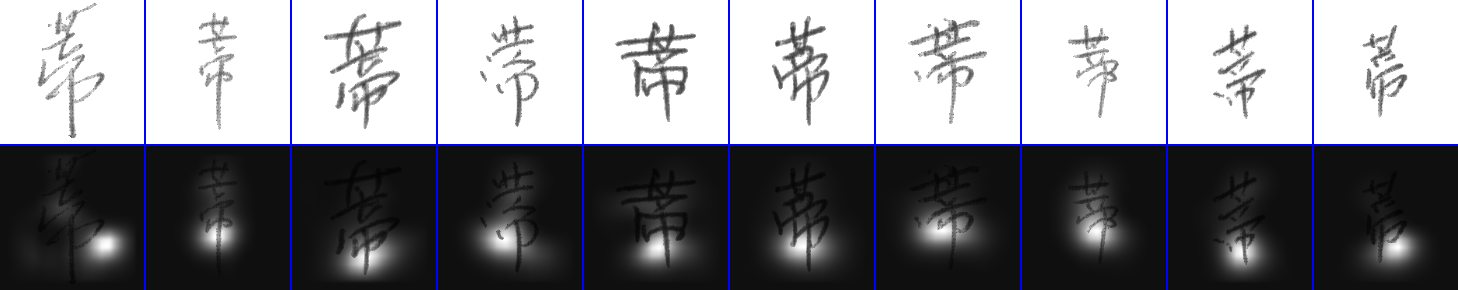}
    }

\caption{Randomly selected images and their corresponding explanations for each class from the Chinese Characters dataset. We can see that the CNN generated explanations tend to point to the same part of the image for each class. The first image for Grass is an interesting failure case. This particular character is quite different from the other instances of the class and as a result the explanation is poor.}
\label{fig:chinese_examples}
\end{figure*}

In Figs. \ref{fig:butterflies}, \ref{fig:oct}, and \ref{fig:chinese_examples} we display some randomly selected images along with their corresponding explanations for each of the three datasets used in the main paper.

